\documentclass[10pt,twocolumn,letterpaper]{article}

\usepackage[pagenumbers]{cvpr} 

\usepackage[dvipsnames]{xcolor}

\definecolor{cvprblue}{rgb}{0.21,0.49,0.74}
\usepackage[pagebackref,breaklinks,colorlinks,citecolor=cvprblue]{hyperref}

\def\paperID{} 
\def\confName{CVPR}
\def\confYear{2024}

\usepackage{graphicx}
\usepackage{amsmath}
\usepackage{amssymb}
\usepackage{lipsum}
\usepackage{booktabs}
\usepackage{xspace}
\usepackage{multirow}
\usepackage[dvipsnames]{xcolor}
\usepackage{float}
\usepackage[numbers]{natbib}
\usepackage{makecell}
\usepackage{xcolor}

\usepackage[capitalize]{cleveref}
\crefname{section}{Sec.}{Secs.}
\Crefname{section}{Section}{Sections}
\Crefname{table}{Table}{Tables}

\usepackage[labelsep=period]{caption}
\RequirePackage{enumitem}
\setlist[itemize]{nosep}

\title{DreamAvatar: Text-and-Shape Guided 3D Human Avatar Generation via Diffusion Models}

\author{Yukang Cao\textsuperscript{1}\thanks{Equal contributions \quad $^\dagger$ Corresponding authors \quad $^\ddagger$ Webpage:  \url{https://yukangcao.github.io/DreamAvatar/}}
\quad
Yan-Pei Cao$^{2*}$
\quad
Kai Han$^{1\dagger}$
\quad
Ying Shan$^{2}$
\quad
Kwan-Yee K. Wong$^{1}$ 
\vspace{0.3em} 
\\
{\normalsize \textsuperscript{1}The University of Hong Kong} \qquad  
{\normalsize \textsuperscript{2}ARC Lab, Tencent PCG}
}

\begin{document}
\maketitle
\begin{abstract}
We present \textbf{DreamAvatar}, a text-and-shape guided framework for generating high-quality 3D human avatars with controllable poses. While encouraging results have been reported by recent methods on text-guided 3D common object generation, generating high-quality human avatars remains an open challenge due to the complexity of the human body's shape, pose, and appearance. We propose DreamAvatar to tackle this challenge, which utilizes a trainable NeRF for predicting density and color for 3D points and pretrained text-to-image diffusion models for providing 2D self-supervision. Specifically, we leverage the SMPL model to provide shape and pose guidance for the generation. We introduce a dual-observation-space design that involves the joint optimization of a canonical space and a posed space that are related by a learnable deformation field. This facilitates the generation of more complete textures and geometry faithful to the target pose.
We also jointly optimize the losses computed from the full body and from the zoomed-in 3D head to alleviate the common multi-face ``Janus'' problem and improve facial details in the generated avatars.
Extensive evaluations demonstrate that DreamAvatar significantly outperforms existing methods, establishing a new state-of-the-art for text-and-shape guided 3D human avatar generation.
\end{abstract}    
\section{Introduction}
\label{sec:intro}

\begin{figure*}[htbp]
  \centering
   \includegraphics[width=1\linewidth]{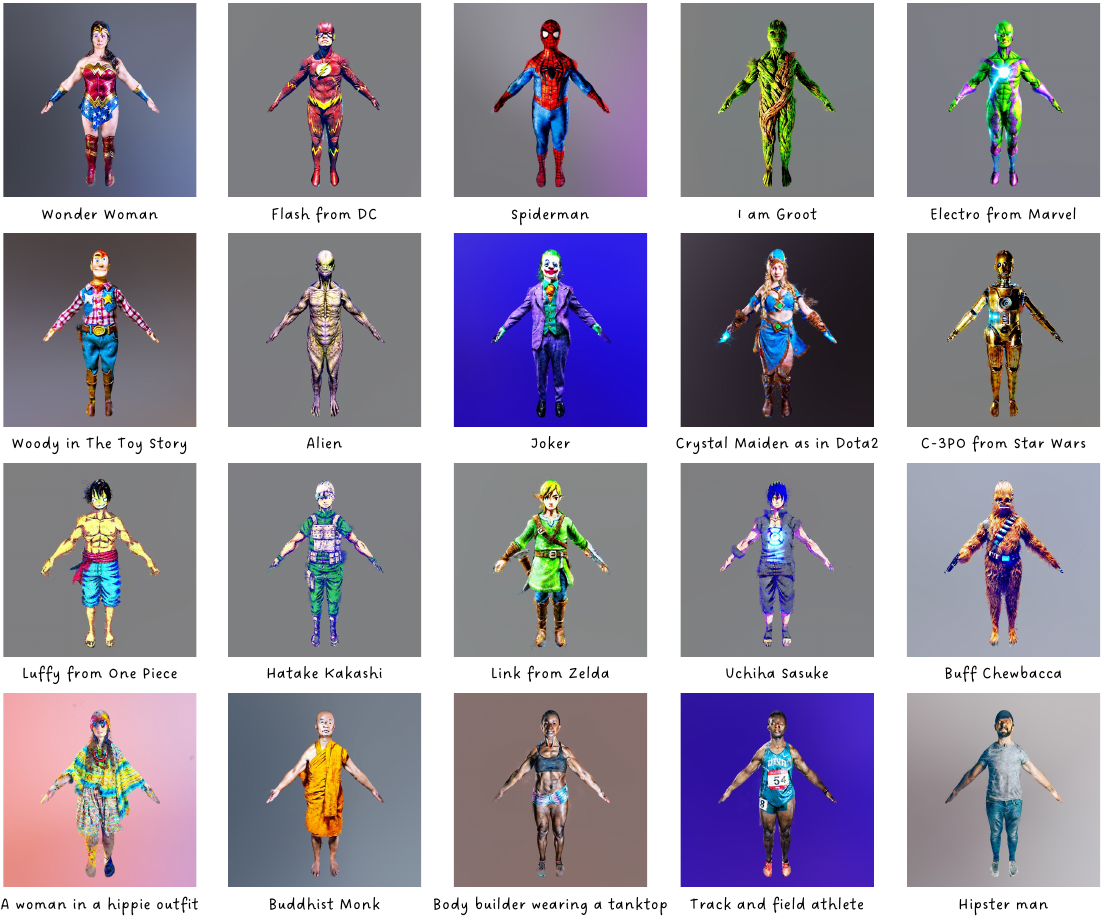}
   \vspace{-1.5em}
  \caption{\textbf{Results of DreamAvatar.} DreamAvatar can generate high-quality geometry and texture for any type of human avatar.}
   \label{fig:Apose}
	\vspace{-1.5em}
   
\end{figure*} 
The creation of 3D graphical human models has received great attention in recent years due to its wide-ranging applications in fields such as film-making, video games, AR/VR, and human-robotic interaction. Traditional methods for building such complex 3D models require thousands of man-hours of trained artists and engineers~\cite{Collet2015HighqualitySF,favalli2012multiview}, making the process both time-consuming and highly expert-dependent. With the development of deep learning methods, we have witnessed the emergence of promising methods~\cite{saito2019pifu, cao2022jiff, xiu2023econ} which can reconstruct 3D human models from monocular images. These techniques, however, still face challenges in fully recovering details from the input images and rely heavily on the training dataset. To tackle these challenges and simplify the modeling process, adopting generative models for 3D human avatar modeling has recently received increasing attention from the research community. This approach has the potential to alleviate the need for large 3D datasets and facilitate easier and more accessible 3D human avatar modeling. 

To leverage the potential of 2D generative image models for 3D content generation, recent methods~\cite{poole2022dreamfusion, lin2022magic3d, metzer2022latent, chen2023fantasia3d, liu2023zero123} have utilized pretrained text-guided image diffusion models to optimize 3D implicit representations (e.g., NeRFs~\cite{mildenhall2020nerf} and DMTet~\cite{munkberg2022nvdiffrec, shen2021dmtet}).  DreamFusion~\cite{poole2022dreamfusion} introduces a novel Score Distillation Sampling (SDS) strategy to self-supervise the optimization process and achieves promising results. 
However, human bodies, which are the primary focus of this paper, exhibit a complex articulated structure, with head, arms, hands, trunk, legs, feet, etc., each capable of posing in various ways. As a result, 
while DreamFusion~\cite{poole2022dreamfusion} and subsequent methods (e.g., Magic3D~\cite{lin2022magic3d}, ProlificDreamer~\cite{wang2023prolificdreamer}, Fantasia3D~\cite{chen2023fantasia3d}) produce impressive results, they lack the proper constraints to enforce consistent 3D human structure and often struggle to generate detailed textures for 3D human avatars. Latent-NeRF~\cite{metzer2022latent} introduces a sketch-shape loss based on the 3D shape guidance, but it still faces challenges in generating reasonable results for human bodies. 

In this paper, we present \emph{\textbf{DreamAvatar}}, a novel framework for generating high-quality 3D human avatars from text prompts and shape priors. 
Inspired by previous works~\cite{poole2022dreamfusion, lin2022magic3d}, DreamAvatar employs a trainable NeRF as the base representation for predicting density and color features for each 3D point. Coupled with pretrained text-to-image diffusion models~\cite{rombach2022high, zhang2023controlnet}, DreamAvatar can be trained to generate 3D avatars using 2D self-supervision. 
\emph{The key innovation of DreamAvatar lies in three main aspects}. \emph{Firstly}, we leverage the SMPL model~\cite{loper2015smpl} to provide a \emph{shape prior}, which yields robust shape and pose guidance for the generation process. \emph{Secondly}, we introduce a \emph{dual-observation-space (DOS) design} 
consisting of a canonical space and a posed space 
that are related by a learnable deformation field and are jointly optimized. This facilitates the generation of more complete textures and geometry faithful to the target pose. 
\emph{Thirdly}, 
we propose to jointly optimize the losses computed from the full body and from the zoomed-in 3D head to alleviate the multi-face ``Janus'' problem and improve facial details in the generated avatars.

\vspace{-0.5em}
We extensively evaluate DreamAvatar on generating movie/anime/video game characters, as well as general people, and compare it with previous methods. Experimental results show that our method significantly outperforms existing methods and can generate high-quality 3D human avatars with text-consistent geometry and geometry-consistent texture. We will make our code publicly available after publication.

\begin{figure*}[htbp]
  \centering
   \includegraphics[width=1\linewidth]{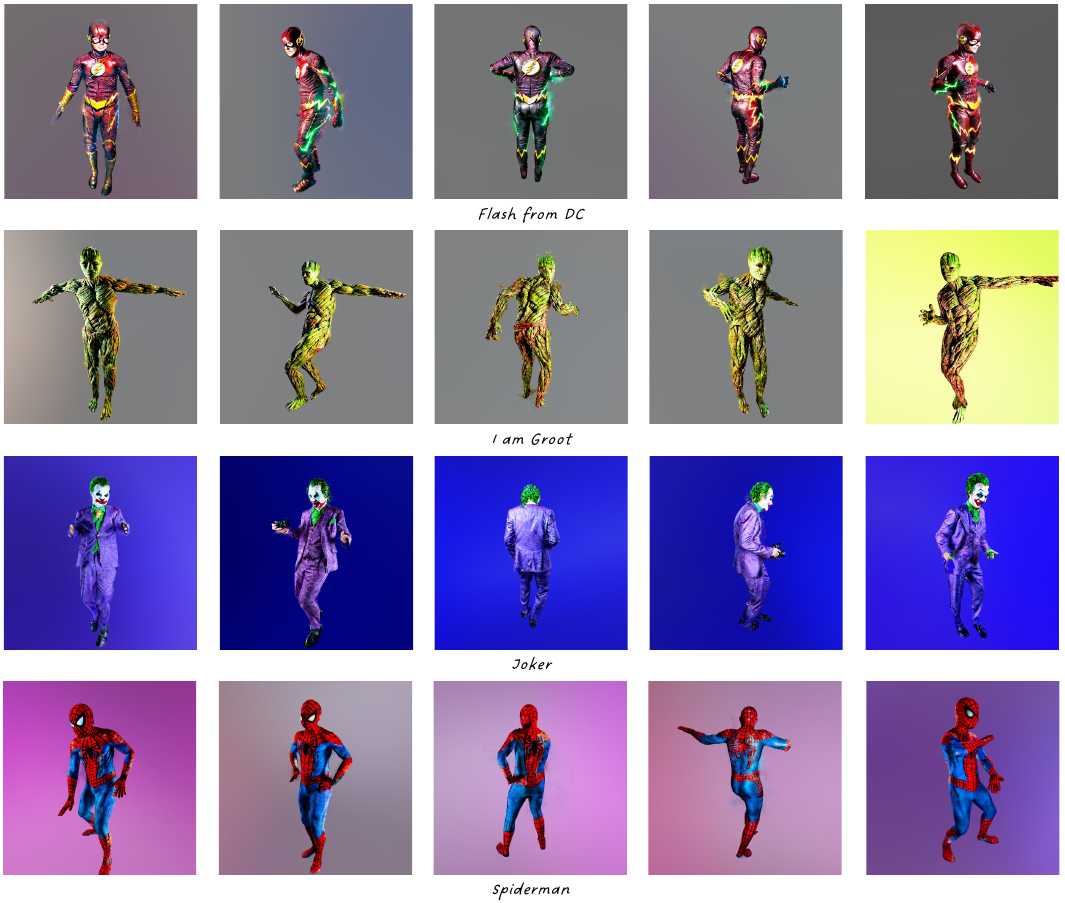}
   \vspace{-2em}
  \caption{\textbf{Avatar generation with different poses.} Our method can handle and control the 3D generation with any pose.}
   \label{fig:poses}
   \vspace{-1.5em}
   
\end{figure*} 
\section{Related Work}
\vspace{-1.2em}
\paragraph{Text-guided 2D image generation.}
Recently, the CLIP model~\cite{radford2021learning} (Contrastive Language-Image Pre-training) was proposed with the aim of classifying images and text by mapping them to a shared feature space. However, this model is not consistent with the way human perceives language, and it may not fully capture the intended meanings. With the improvements in neural networks and text-image datasets, the diffusion model has been introduced to handle more complex semantic concepts~\cite{balaji2022ediffi, metzer2022latent, ramesh2022hierarchical, saharia2022photorealistic}. Follow-up methods are designed to improve computational efficiency, for instance, through utilizing a cascade of super-resolution models~\cite{balaji2022ediffi, saharia2022photorealistic} or sampling from a low-resolution latent space and decoding the latent features into high-resolution images~\cite{metzer2022latent}. DreamBooth~\cite{ruiz2022dreambooth} fine-tunes the diffusion model for certain subjects, while ControlNet~\cite{zhang2023adding} and T2I-Adapter~\cite{mou2023t2i} propose controlling the pretrained diffusion models with additional information. However, text-to-3D generation remains a challenge due to the lack of text-3D paired datasets and the associated high training cost .

\vspace{-1.2em}
\paragraph{Text-guided 3D content generation.}
Text-guided 3D content generation methods have emerged based on the success of text-guided 2D image generation. Earlier works, such as CLIP-forge~\cite{sanghi2022clip}, generate objects by learning a normalizing flow model from textual descriptions, but these methods are computationally expensive. DreamField~\cite{jain2021dreamfields}, CLIP-mesh~\cite{khalid2022clipmesh}, AvatarCLIP~\cite{hong2022avatarclip}, Text2Mesh~\cite{Michel_2022_CVPR}, and Dream3D~\cite{xu2022dream3d} rely on a pretrained image-text model~\cite{radford2021learning} to optimize the underlying 3D representation (e.g., NeRF or mesh). 

Recently, DreamFusion~\cite{poole2022dreamfusion} proposes score distillation sampling based on the pretrained diffusion model~\cite{saharia2022photorealistic} to enable text-guided 3D generation. Magic3D~\cite{lin2022magic3d} improves it by introducing a coarse-to-fine pipeline to generate fine-grained 3D textured meshes. Point-E~\cite{nichol2022point} and Shap-E~\cite{jun2023shap} optimize the point cloud based on the diffusion model. Latent-NeRF~\cite{metzer2022latent} improves training efficiency by directly optimizing the latent features. TEXTure~\cite{richardson2023texture} applies a depth-diffusion model~\cite{rombach2022high} to generate texture maps for a given 3D mesh. 
Fantasia3D~\cite{chen2023fantasia3d} proposes a disentangled training strategy for geometry and texture generation.
Guide3D~\cite{cao2023guide3d} proposes to transfer multi-view generated images to 3D avatars.
ProlificDreamer~\cite{wang2023prolificdreamer} introduces Variational Score Distillation (VSD) for better diversity and quality.
Despite their promising performance, these methods still struggle to generate text-guided 3D human avatars due to the inherent challenges of this task.
\vspace{-1.2em}

\paragraph{3D human generative models.}
3D generative methods based on 3D voxel grids~\cite{gadelha20173d, henzler2019escaping, lunz2020inverse, wu2016learning}, point clouds~\cite{achlioptas2018learning, luo2021diffusion, mo2019structurenet, yang2019pointflow, zeng2022lion, zhou20213d}, and meshes~\cite{zhang2020image} 
often require expensive and limited 3D datasets. In recent years, various methods~\cite{2021narf, liu2021neural, peng2021neural, weng2022humannerf,chen2019learning} have been proposed to utilize NeRF and train on 2D human videos for novel view synthesis. Following these works, EG3D~\cite{Chan2022} and GNARF~\cite{bergman2022gnarf} propose a tri-plane representation and use GANs for 3D generation from latent codes. ENARF-GAN~\cite{noguchi2022unsupervised} extends NARF~\cite{2021narf} to human representation. Meanwhile, EVA3D~\cite{hong2023evad} and HumanGen~\cite{jiang2022humangen} propose to generate human radiance fields directly from the 2D StyleGAN-Human~\cite{fu2022stylegan} dataset. Although these methods have produced convincing results, they do not have the ability to ``dream" or generate new subjects that have not been seen during training.
\section{Methodology}

\begin{figure*}[htbp]
  \centering
   \includegraphics[width=1\linewidth]{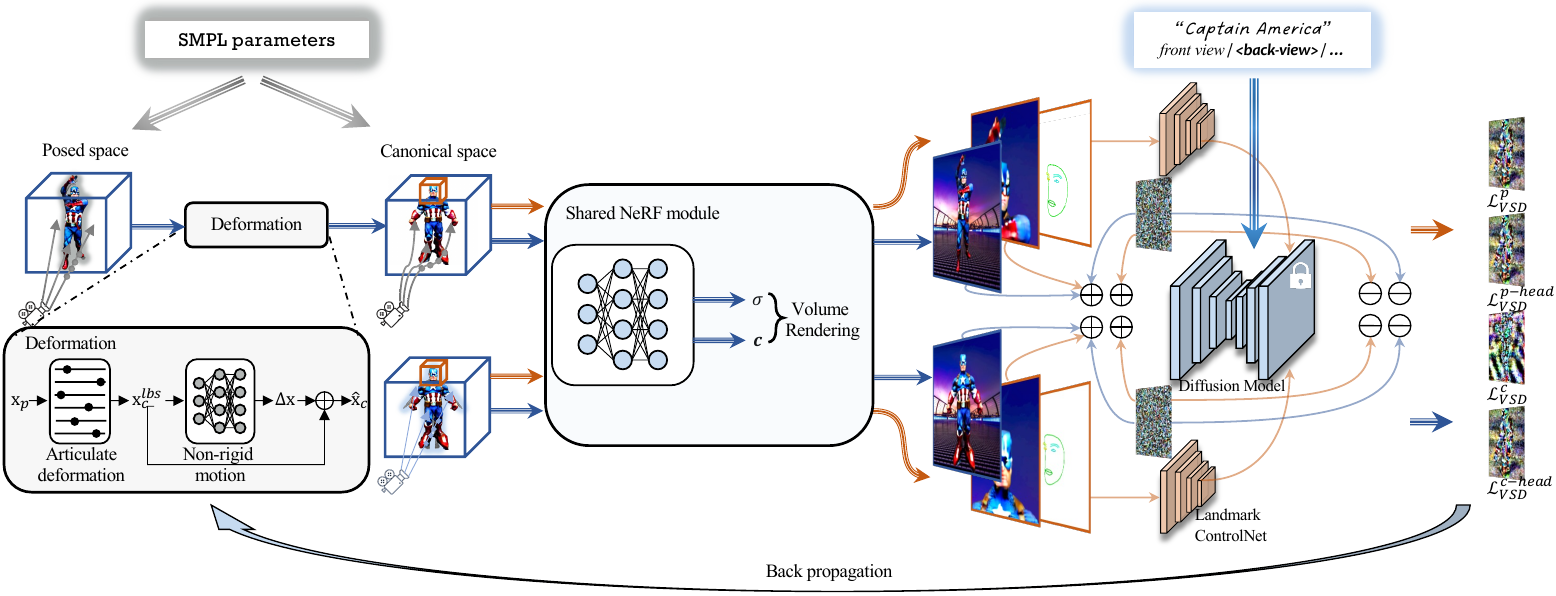}
  \caption{\textbf{Overview of DreamAvatar.} Our network takes as input a text prompt and SMPL parameters to optimize a trainable NeRF via a pretrained denoising stable diffusion model. At the core of our network are two observation spaces, namely the canonical space and the posed space, that are related by an SMPL-based learnable deformation field. This dual-observation-space design facilities the generation of more complete textures and geometry faithful to the target pose.}
   \label{fig:pipeline}
    \vspace{-1em}
\end{figure*} 
Here, we introduce our text-and-shape guided generative network, DreamAvatar, which utilizes a trainable NeRF and pretrained diffusion models~\cite{rombach2022high, zhang2023controlnet} to generate 3D human avatars under controllable poses. DreamAvatar incorporates two observation spaces, namely a canonical space and a posed space, which are related by a learnable deformation field and are jointly optimized through a shared trainable NeRF (see \cref{fig:pipeline}). 
We jointly optimize the losses computed from the full-body and from the zoom-in 3D head to alleviate the multi-face ``Janus'' problem and improve the facial details in the generated avatars.
In the following subsections, we first provide the preliminaries that underpin our method in \cref{sec:preliminary}. Next, we delve into the details of our method and discuss: (1) the density field derived from the SMPL model for evolving the geometry, (2) the dual observation spaces related by a learnable deformation field, and (3) 
the joint optimization of the losses computed from the full body and from the zoom-in 3D head
in \cref{sec:method}. 

\subsection{Preliminaries}
\label{sec:preliminary}

\paragraph{Text-guided 3D generation methods} Recent text-guided 3D generation models~\cite{poole2022dreamfusion, wang2023prolificdreamer} showcase promising results by incorporating three fundamental components: 

(1) \textit{NeRF} that represents a 3D scene via an implicit function, formulated as
\begin{equation}
    F_{\theta}(\gamma(\mathbf{x})) \mapsto (\sigma, \mathbf{c}),
\end{equation}
where $\mathbf{x}$ is a 3D point that is processed by the grid frequency encoder $\gamma(\cdot)$~\cite{mildenhall2020nerf}, and $\sigma$ and $\mathbf{c}$ denote its density value and color respectively. Typically, the implicit function $F_\theta(\cdot)$ is implemented as an MLP with trainable parameters $\theta$. 

(2) \textit{Volume Rendering} technique that effectively renders a 3D scene onto a 2D image. For each image pixel, the rendering is done by casting a ray $\mathbf{r}$ from the pixel location into the 3D scene and sampling 3D points $\boldsymbol{\mu}_i$ along $\mathbf{r}$. The density and color of the sampled points are predicted by $F_{\theta}$. The RGB color $C$ of each image pixel is then given by
\begin{equation}
    \label{eq:volume_rendering}
    C(\mathbf{r}) = \sum_{i}W_i\mathbf{c}_i, \quad W_i = \alpha_i \prod_{j < i}(1-\alpha_j)
\end{equation}
where $\alpha_i = 1 - e^{(-\sigma_i ||\boldsymbol{\mu}_i - \boldsymbol{\mu}_{i+1}||)}$. 

(3) \textit{Variational Score Distillation (VSD)} derived on text-guided diffusion models $\phi$~\cite{rombach2022high, saharia2022photorealistic}. We employ a pretrained diffusion model~\cite{rombach2022high, zhang2023controlnet} with a learned denoising function $\epsilon_{\text{pre}}(\boldsymbol{x_t};y,t)$. Here $\boldsymbol{x}_t$ denotes the noisy image at timestep $t$, and $y$ is the text embedding. 
Given an image $\boldsymbol{g}(\theta,c)$ rendered from the NeRF with camera parameters $c$, we add random noise $\epsilon$ to obtain a noisy image $\boldsymbol{x}_t=\alpha_t\boldsymbol{g}(\theta,c)+\sigma_t\epsilon$ ($\alpha_t$ and $\sigma_t$ are hyperparameters). We then parameterize $\epsilon$ using LoRA (Low-Rank Adaptation~\citep{hu2021lora,diffusion-lora}) of the pretrained model $\epsilon_{\text{pre}}(x_t;y,t)$ to obtain $\epsilon_\phi$, and add camera parameters $c$ to the condition embeddings in the network. The gradient for updating the NeRF is then given by

\begin{equation}
    \label{eq:vds}
    \small
    \footnotesize
    \nabla_{\theta}\mathcal{L}_{\text{VSD}}(\theta) \triangleq \mathbb{E}_{t,\epsilon,c}\left[
    \omega(t)
    \left( \epsilon_{\text{pre}}(\boldsymbol{x}_t,t,y)\!-\!\epsilon_{\phi}(\boldsymbol{x}_t,t,c,y)  \right)
    {\scriptscriptstyle \frac{\partial \boldsymbol{g}(\theta,c)}{\partial \theta}}\right],
\end{equation}
where $w(t)$ is a weighting function that depends on the timestep $t$.

\vspace{-1em}
\paragraph{SMPL~\cite{bogo2016keep, pavlakos2019expressive} 3D parametric human model} It builds a 3D human shape using 6,890 body vertices. Formally, by assembling pose parameters $\xi$ and shape parameters $\beta$, we can obtain the 3D SMPL human model by: 
\begin{eqnarray}\label{eq:SMPLX}
    \small
    T_P(\beta, \xi)&=&\Bar{T} + B_S(\beta; \mathcal{S}) + B_P(\xi; \mathcal{P}), \\ 
    M(\beta, \xi)&=&\mathtt{LBS}(T_P(\beta, \xi), J(\beta), \xi, \mathcal{W}), 
\end{eqnarray}
where $T_P(\cdot)$ represents the non-rigid deformation from the canonical model $\Bar{T}$ using the shape blend shape function $B_S$ and pose blend shape function $B_P$. $\mathcal{S}$ and $\mathcal{P}$ are the principal components of vertex displacements. $\mathtt{LBS}(\cdot)$ denotes the linear blend skinning function, corresponding to articulated deformation. It poses $T_P(\cdot)$ based on the pose parameters $\xi$ and joint locations $J(\beta)$, using the blend weights $\mathcal{W}$, individually for each body vertex:
\begin{equation}\label{eq:deformation}
    \small
    \mathbf{v}_p = \mathcal{G} \cdot \mathbf{v}_c, \quad \mathcal{G} = \sum_{k=1}^{K} w_k \mathcal{G}_k (\xi, {j}_k),
\end{equation}
where $\mathbf{v}_c$ is an SMPL vertex under the canonical pose, $\mathbf{v}_p$ denotes the corresponding vertex under the given pose, $w_k$ is the skinning weight, $\mathcal{G}_k (\xi, j_k)$ is the affine deformation that transforms the $k$-th joint ${j}_k$ from the canonical space to posed space, and $K$ is the number of neighboring joints. 

Unlike the original SMPL which defines ``T-pose" as the canonical model, here, we adopt ``A-pose" as the canonical model which is a more natural human rest pose for the diffusion models to understand (see \cref{fig:pipeline}).

\subsection{DreamAvatar}
\label{sec:method}

As illustrated in \cref{fig:pipeline}, our proposed framework takes as input a text prompt and SMPL parameters, defining the target shape and pose in the posed space. DreamAvatar conducts Variational Score Distillation (VSD)-based optimization~\cite{wang2023prolificdreamer} in both the canonical space and posed space simultaneously and learns a deformation field relating the two spaces. To represent the canonical space and posed space, we utilize an extended neural radiance field where the density, RGB color value, and normal direction of each sample point can be queried and optimized. We utilize the input SMPL parameters to handle different body parts separately and derive good initial density values in each space. 
To alleviate the multi-face ``Janus'' problem and improve facial details in the generated avatars, we optimize, in addition to the loss computed from the full body, the loss computed from the zoomed-in 3D head using a landmark-based ControlNet~\cite{zhang2023adding} and a learned special $<$back-view$>$ token~\cite{gal2022textual-inversion}.

\vspace{-1em}
\paragraph{SMPL-derived density fields} 
We propose to make our NeRF evolve from the density field derived from the input SMPL model. Specifically, given a 3D point $\mathbf{x}_c$ in the canonical space, we first calculate its signed distance $d_c$ to the SMPL surface in the canonical pose and convert it to a density value $\bar{\sigma}_{c}$ using
\begin{equation*}
    \small
    \bar{\sigma}_{c} = \mathtt{max}(0, \mathtt{softplus}^{-1}(\frac{1}{a}\mathtt{sigmoid}(-d_c/a))),  
\end{equation*}
where $\mathtt{sigmoid}(x) = 1 / (1 + e^{-x})$, $\mathtt{softplus}^{-1}(x) = \mathtt{log}(e^x-1)$, and $a$ is a predefined hyperparameter~\cite{xu2022dream3d} which is set to 0.001 in our experiments. 
Similarly, given a point $\mathbf{x}_p$ in the posed space (the upper branch in \cref{fig:pipeline}), we compute its density value $\bar{\sigma}_{p}$ from its signed distance $d_p$ to the SMPL surface in the target pose.
\vspace{-1em}
\paragraph{Deformation field} Inspired by HumanNeRF~\cite{weng2022humannerf}, we employ a deformation field to map points $\mathbf{x}_p$ from the posed space to their corresponding points $\hat{\mathbf{x}}_c$ in the canonical space. The deformation field is composed of two parts, namely (1) articulated deformation that applies the inverse transformation of SMPL linear blend skinning $\mathtt{LBS}(\cdot)$ (\cref{eq:SMPLX}), and (2) non-rigid motion modelled by an MLP to learn the corrective offset: 
\begin{equation}
    \small
    \hat{\mathbf{x}}_c = \mathcal{G}^{-1} \cdot \mathbf{x}_p + \mathtt{MLP}_{\theta_{\text{NR}}}(\gamma(\mathcal{G}^{-1} \cdot \mathbf{x}_p)),
\end{equation}
where $\mathcal{G}$ is obtained from posed SMPL vertex closest to $\mathbf{x}_p$.

\begin{figure*}[htbp]
  \centering
   \includegraphics[width=1\linewidth]{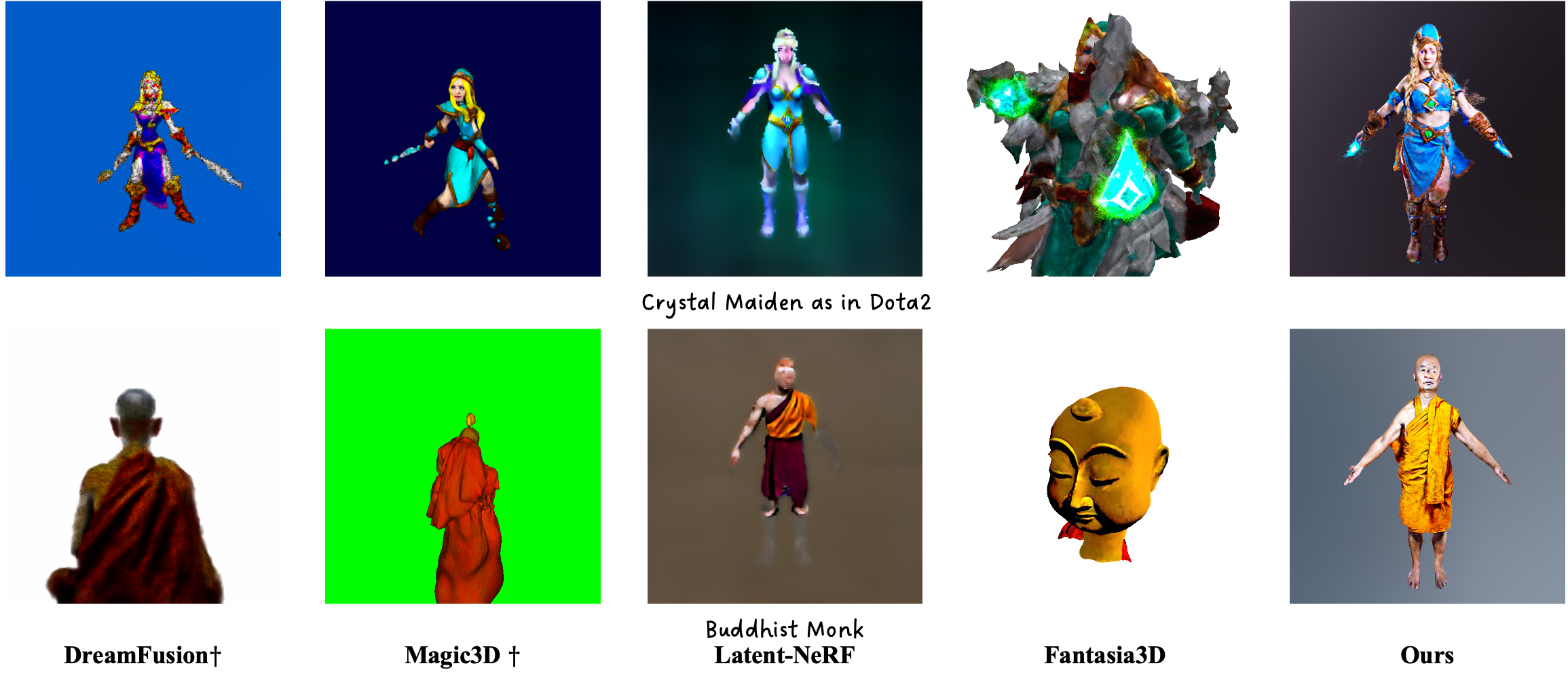}
    \vspace{-2em}
  \caption{\textbf{Comparison with existing text-to-3D baselines.} While baselines struggle to ensure structurally and topologically correct geometry, our method can produce high-quality geometry and texture.}
    \vspace{-1em}
   \label{fig:comparison1}   
\end{figure*} 

\begin{figure*}[htbp]
  \centering
   \includegraphics[width=1\linewidth]{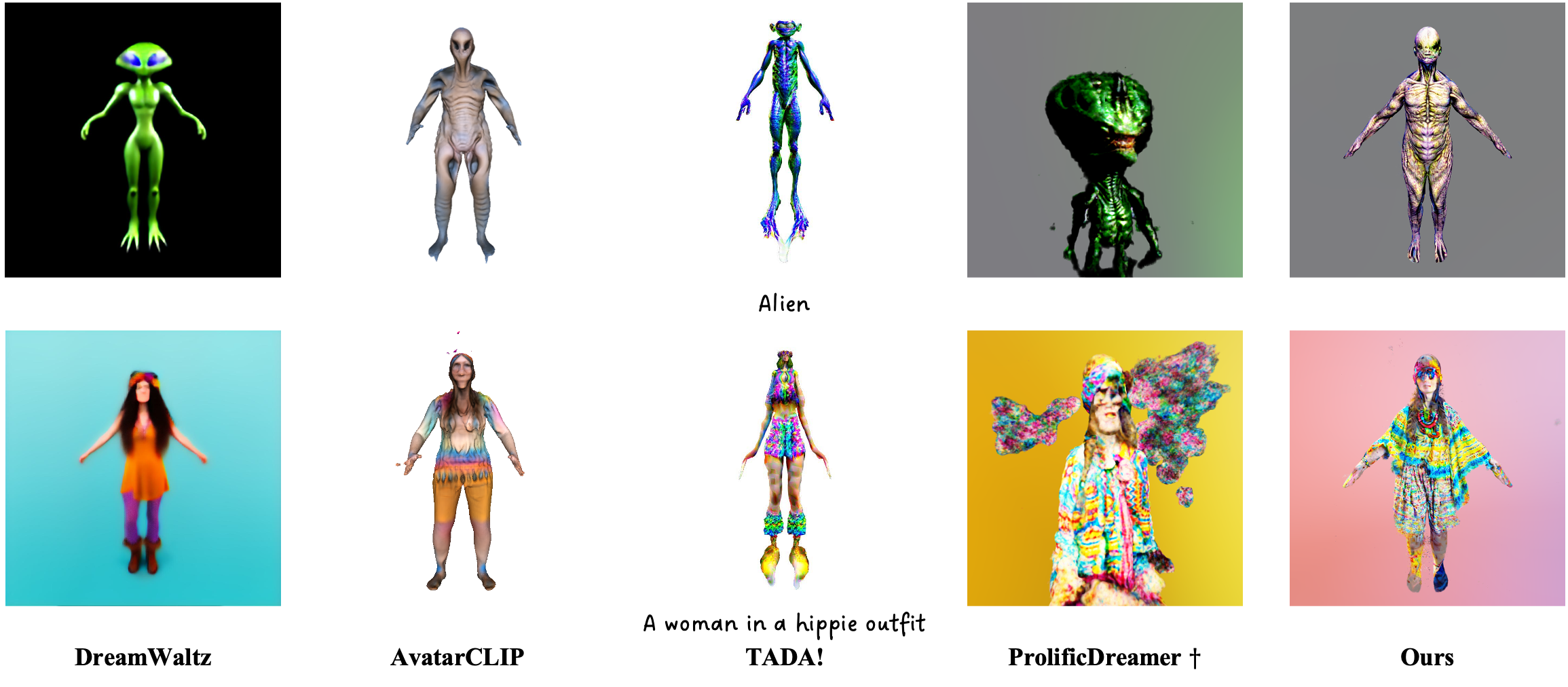}
    \vspace{-2em}
  \caption{\textbf{Comparison with existing avatar-specified baselines.} Our method can generate 3D human avatars with geometry and texture with much higher resolution.}
   \label{fig:comparison2}
    \vspace{-1.5em}
\end{figure*} 
\vspace{-1em}
\paragraph{Dual observation spaces (DOS)}
Given a 3D point $\mathbf{x}_c$ in the canonical space and $\mathbf{x}_p$ in the posed space, we compute their density values $\sigma_c, \sigma_p$ and latent color features $\mathbf{c}_c, \mathbf{c}_p$ by
\begin{equation}
    \small
    F(\mathbf{x}_c, \bar{\sigma}_{c}) = F_{\theta}(\gamma(\mathbf{x}_c)) + (\bar{\sigma}_{c}, \mathbf{0})  \mapsto (\sigma_c, \mathbf{c}_c),
\end{equation}
\begin{equation}
    \small
    F(\mathbf{x}_p, \bar{\sigma}_{p}) = F_{\theta}(\gamma(\hat{\mathbf{x}}_c)) + (\bar{\sigma}_{p}, \mathbf{0})  \mapsto (\sigma_p, \mathbf{c}_p), 
\end{equation}
where $\hat{\mathbf{x}}_c$ denotes the corresponding point of $\mathbf{x}_p$ in the canonical space, and $\bar{\sigma}_{c}$ and $\bar{\sigma}_{p}$ are the SMPL-derived density values in the canonical space and posed space respectively.
Our DOS design serves two main purposes. Firstly, the avatar in rest pose in the canonical space exhibits minimum self-occlusion. Observations in the canonical space therefore facilitate the generation of more complete textures. Observations in the posed space, on the other hand, facilitate the generation of geometry faithful to the target pose. They also provide extra supervision in optimizing the NeRF. 
Our DOS design also differentiates itself from MPS-NeRF~\cite{gao2022mps} and TADA!~\cite{liao2023tada} by prioritizing joint optimization and mutual distillation between the canonical space and posed space. In contrast, existing methods only utilize observations in the posed space without fully exploiting information in the canonical space.

\begin{figure*}[htbp]
  \centering
   \includegraphics[width=1\linewidth]{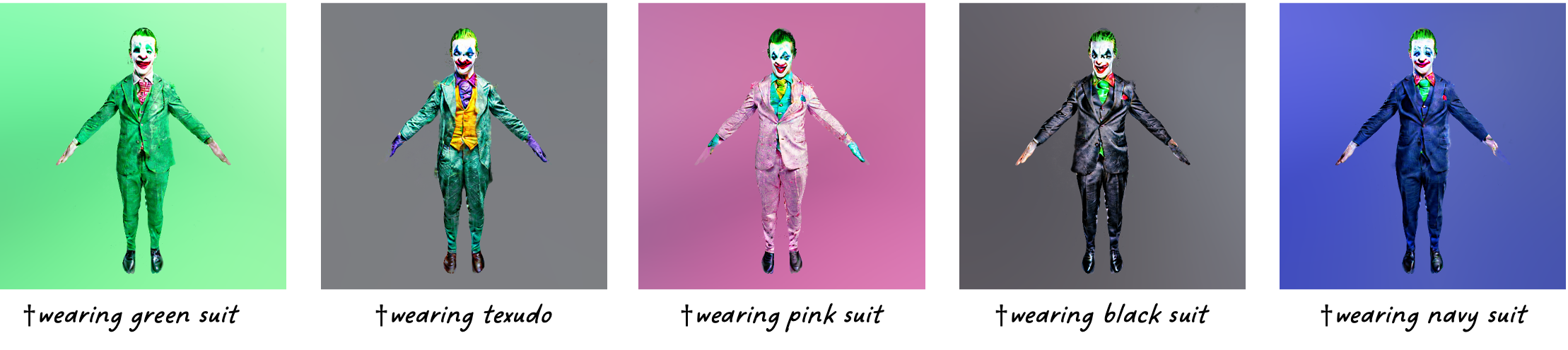}
  \vspace{-2em}
  \caption{\textbf{Text manipulation over the avatar generations.} $\dagger$ indicates ``Joker''}
   \label{fig:styles}
   \vspace{-0.5em}
   
\end{figure*} 
\begin{figure*}[htbp]
  \centering
   \includegraphics[width=1\linewidth]{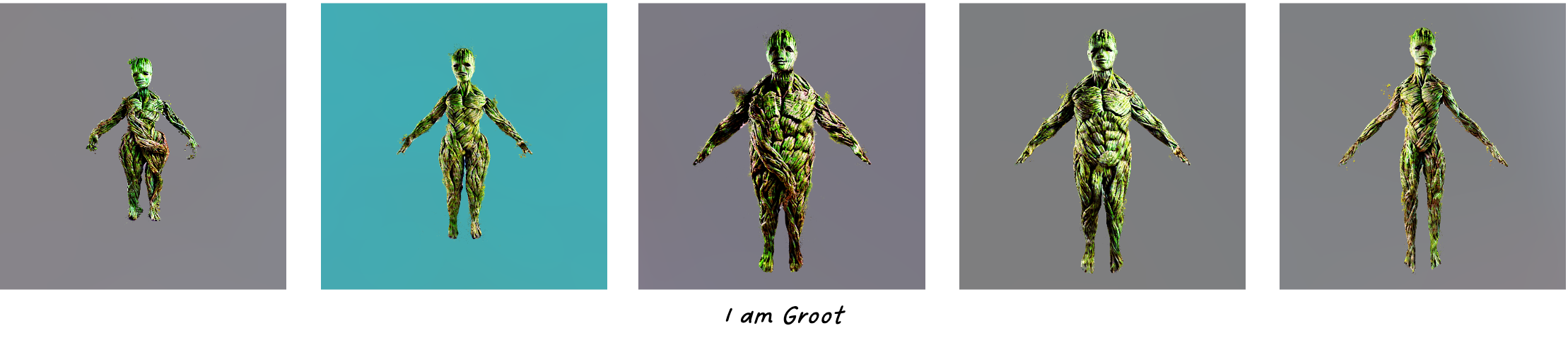}
  \vspace{-2em}
  \caption{\textbf{Shape modification via SMPL shape parameters.} Our method can generate 3D avatars with different shapes based on the input SMPL shape parameters, e.g., ``short'', ``fat'', ``thin'', ``tall'' as above from left to right.}
   \label{fig:shape}
   \vspace{-1em}
\end{figure*} 
\begin{figure*}[t]
  \centering
   \includegraphics[width=1\linewidth]{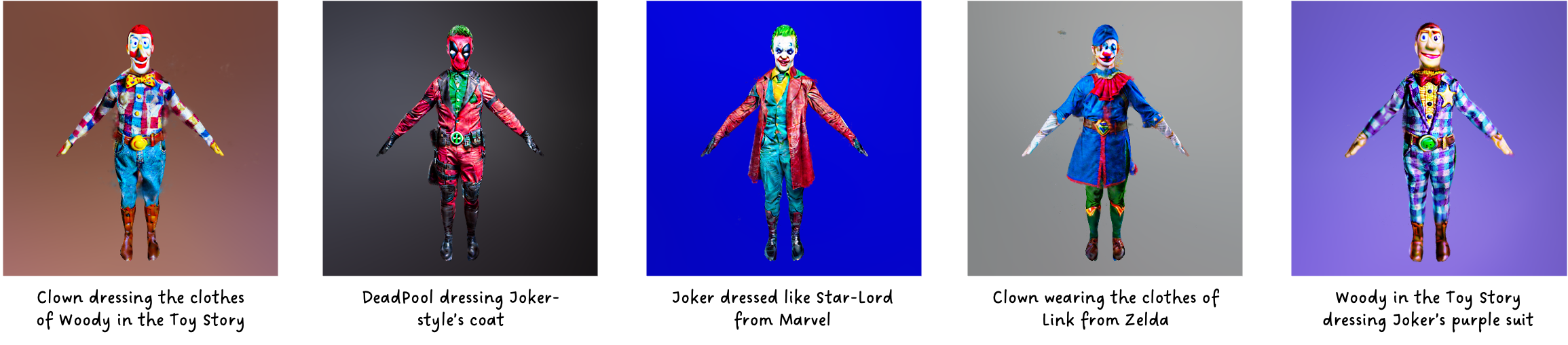}
  \vspace{-1em}
  \caption{\textbf{Attributes integration between head and body parts.}}
   \label{fig:combine}
   \vspace{-1em}
   
\end{figure*} 
\vspace{-1em}
\paragraph{Zoomed-in head} 
Inspired by our previous work~\cite{cao2022jiff} which enhances human reconstruction by learning to recover details in the zoomed-in face, we propose to optimize, in addition to the loss computed from the full body, the loss computed from the zoomed-in 3D head to improve the facial details in the generated 3D avatars. Specifically, we render a zoomed-in head image at each iteration for computing the VSD loss on head. To alleviate the common multi-face ``Janus'' problem, we follow our previous work~\cite{han2023headsculpt} to employ a landmark-based ControlNet $\mathcal{C}$~\cite{zhang2023adding} 
and a learned special $<$back-view$>$ token~\cite{gal2022textual-inversion} to compute the head VSD loss.
The gradient for updating the NeRF now becomes

\vspace{-1em}
\begin{equation}
    \small
    \nabla_{\theta}\mathcal{L}_{\text{VSD}}^{\text{head}}(\theta) \triangleq \mathbb{E}_{t,\epsilon,c}\left[
    \omega(t)
    \left( \epsilon_{\text{pre}}(\cdot) - \epsilon_{\phi}(\cdot)  \right)
    \frac{\partial \boldsymbol{g}(\theta,c)}{\partial \theta}\right],
\end{equation}
\begin{equation*}
    \small
    \epsilon_{\text{pre}}(\cdot) = \epsilon_{\text{pre}}(\boldsymbol{x}_t,t,y, \mathcal{C}(\mathcal{P}_{\pi})), \! 
    \epsilon_{\phi}(\cdot) = \epsilon_{\phi}(\boldsymbol{x}_t,t,c,y, \mathcal{C}(\mathcal{P}_{\pi})),
\end{equation*}
where $\mathcal{P}_{\pi}$ denotes the facial landmark map obtained from the projection of the SMPL head model.

\vspace{-1em}
\section{Experiments}
\label{sec:experiments}
We now validate the effectiveness and capability of our proposed framework on a variety of text prompts and provide comparisons with existing text-guided 3D generation methods using the same text prompts.

\vspace{-1.5em}
\paragraph{Implementation details}
We follow threestudio~\cite{threestudio2023} to implement the NeRF~\cite{mildenhall2020nerf} and Variational Score Distillation in our method. We utilize Ver-2.1 of Stable Diffusion~\cite{stable-diffusion} and Ver-1.1 of ControlNet~\cite{zhang2023controlnet, ControlNetMediaPipeFace} in our implementation. Typically, for each text prompt, we train our network for $10,000$ iterations on one single NVIDIA A40 GPU.

\vspace{-1.5em}
\paragraph{Baseline methods} 
We compare our method with 
DreamFusion~\cite{poole2022dreamfusion}, Latent-NeRF~\cite{metzer2022latent}, Fantasia3D~\cite{chen2023fantasia3d}, Magic3D~\cite{lin2022magic3d},  ProlificDreamer~\cite{wang2023prolificdreamer},
DreamWaltz~\cite{huang2023dreamwaltz}, AvatarCLIP~\cite{hong2022avatarclip}, and TADA!~\cite{liao2023tada}. We are not able to compare with AvatarBooth~\cite{zeng2023AvatarBoothHA} and AvatarVerse~\cite{zhang2023avatarverse} as their codes are not publicly available.

\subsection{Qualitative Evaluations}
\paragraph{Avatar generation with different styles} In \cref{fig:Apose}, we provide a diverse set of 3D human avatars, \eg, real-world human beings, movie, anime, and video game characters, generated by our DreamAvatar. We can consistently observe high-quality geometry and texture from all these examples. 

\vspace{-1.5em}
\paragraph{Avatar generation under different poses} In \cref{fig:poses}, we validate the effectiveness of our method for generating 3D human avatars in various poses, which is not achievable by other existing methods due to the absence of the shape prior. Thanks to our DOS design, DreamAvatar can maintain high-quality texture and geometry for extreme poses, \eg, complex poses with severe self-occlusion. 

\vspace{-1.5em}
\paragraph{Comparison with SOTA methods} We provide qualitative comparisons with existing SOTA methods in \cref{fig:comparison1} and \cref{fig:comparison2}. 
We can observe that our method consistently achieves topologically and structurally correct geometry and texture compared to baseline methods, and outperforms the avatar-specified generative methods with much better and higher-resolution texture and geometry. See supplementary for more comparisons.

\vspace{-1.5em}
\paragraph{Text manipulation on avatar generation} We explore the capabilities of DreamAvatar by editing the text prompt for controlled generation (see~\cref{fig:styles}). 
Our method can generate faithful avatars that accurately embody the provided text, incorporating additional descriptive information and capturing the unique characteristics of the main subject.

\vspace{-1.5em}
\paragraph{Shape modification via SMPL shape parameters} We further demonstrate the possibility of our method to generate different sizes of 3D human avatars, \eg, thin, short, tall, fat, by editing the SMPL shape parameters in \cref{fig:shape}.

\begin{figure}[h]
  \centering
   \includegraphics[width=1\linewidth]{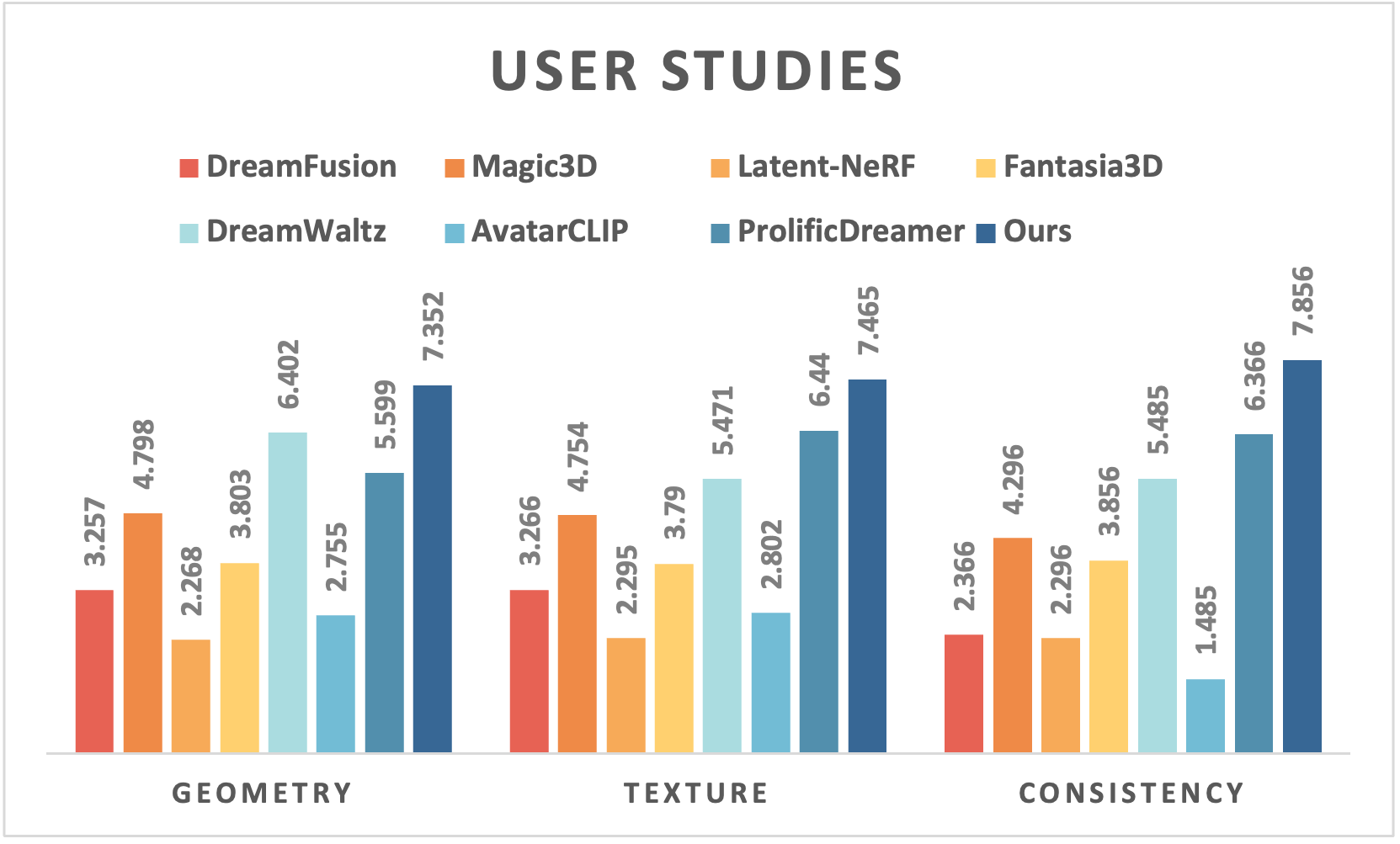}
  \caption{\textbf{User studies on rotating 3D human avatars.}}
   \label{fig:user_study}
   \vspace{-2em}
   
\end{figure} 
\vspace{-1.5em}
\paragraph{Attributes integration: Zoomed-in head and full body} 
Benefiting from the joint modeling of the zoomed-in 3D head and full body, our method seamlessly integrates the unique attributes derived from both head and body characters. See visualizations in \cref{fig:combine}. In contrast to the conventional approach of separately modeling head and body parts, DreamAvatar harmoniously combines these elements, resulting in a cohesive model that faithfully captures the essence of both subjects.

\subsection{User Studies}
We conduct user studies to compare with SOTA methods~\cite{poole2022dreamfusion, lin2022magic3d, metzer2022latent, chen2023fantasia3d, wang2023prolificdreamer, hong2022avatarclip, huang2023dreamwaltz}. 25 volunteers are presented with 20 examples to rate these methods in terms of (1) geometry quality, (2) texture quality, and (3) consistency with the text from 1 (worst) to 8 (best). The final rates in \cref{fig:user_study} clearly show that our method achieves the best rankings in all three aspects.

\subsection{Further Analysis}
\begin{figure}[htbp]
  \centering
   \includegraphics[width=1\linewidth]{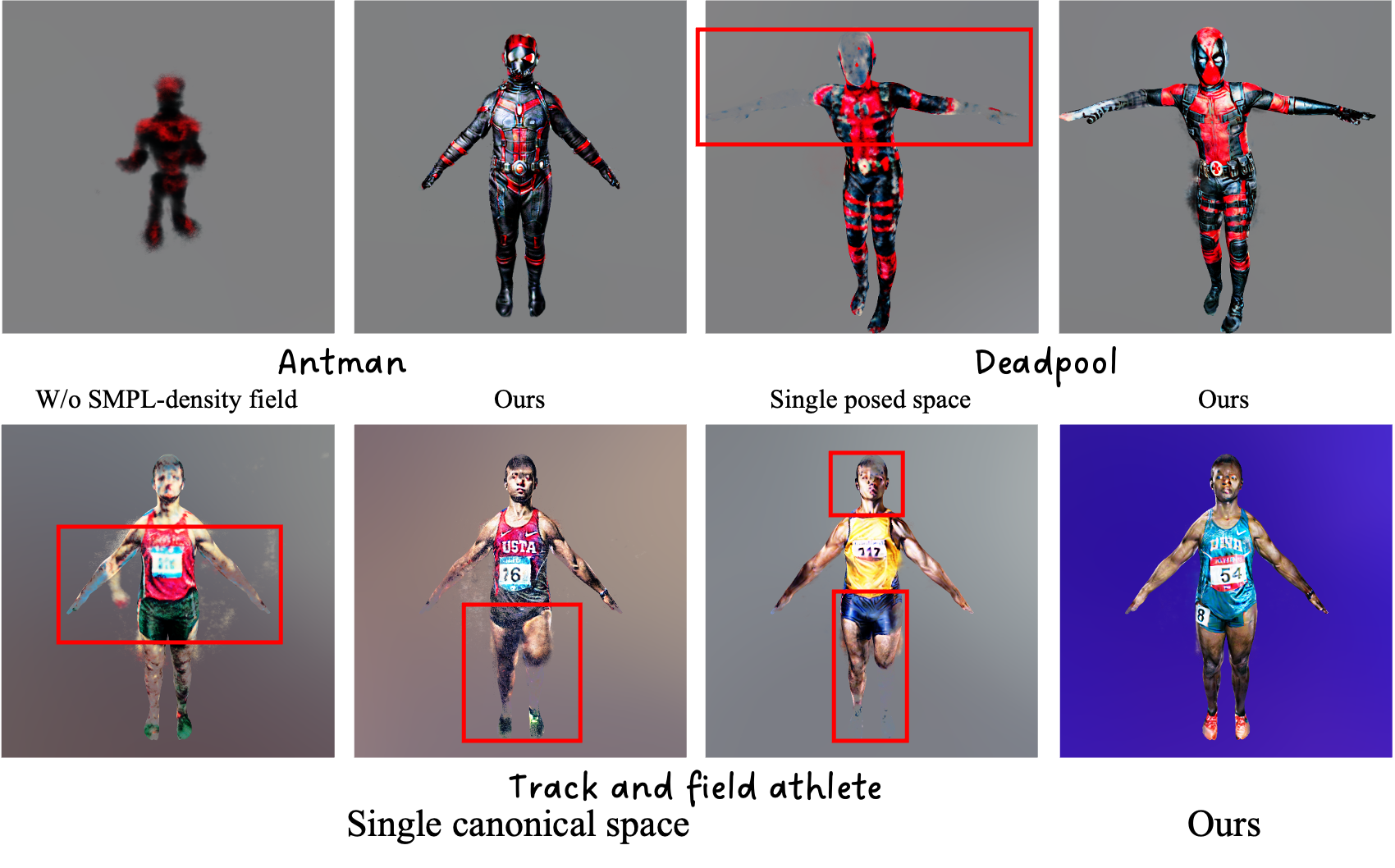}
   \vspace{-1.5em}
  \caption{\textbf{Analysis of the setup for SMPL-derived density $\bar{\sigma}_c, \bar{\sigma}_p$, and dual-space design.}}
  \vspace{-1.em}
   \label{fig:ablation_2}
   
\end{figure}

\paragraph{Effectiveness of SMPL-derived density}
We optimize the NeRF without using the SMPL-derived densities $\bar{\sigma}_c, \bar{\sigma}_p$ as the basis for density prediction. We find that without using $\bar{\sigma}_c$ and $\bar{\sigma}_p$, (1) the generated avatars exhibit low-quality geometry and texture with strong outliers, and (2) the generated shapes are not constrained to reasonable human bodies and are not view-consistent (see ``w/o SMPL-density field'' in \cref{fig:ablation_2}).  

\begin{figure}[htbp]
  \centering
   \includegraphics[width=1\linewidth]{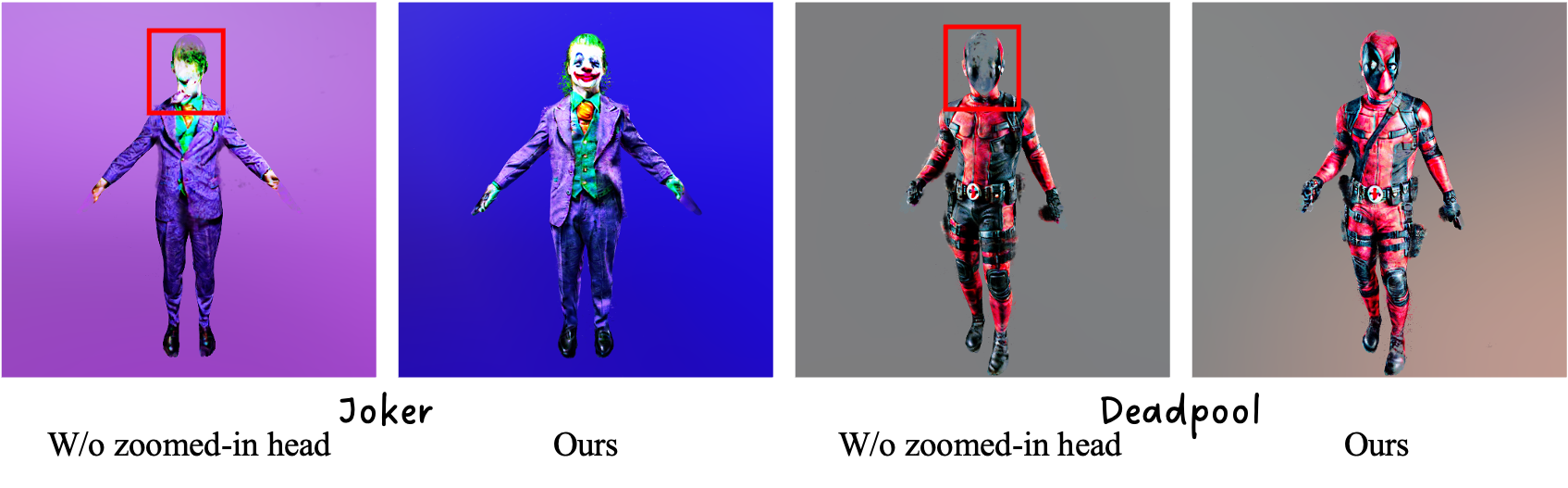}
   \vspace{-1.5em}
  \caption{\textbf{Analysis of joint modeling of head region.}}
  \vspace{-1.8em}
   \label{fig:ablation_head}
   
\end{figure}

\vspace{-1.5em}
\paragraph{Effectiveness of incorporating head VSD loss}
In order to assess its impact, we conduct ablation studies by disabling the head VSD loss. The results are presented in \cref{fig:ablation_head}. Notably, we observe a significant drop in the quality of the generated head region. Moreover, the multi-face "Janus" problem becomes more pronounced in the generated content.

\vspace{-1.5em}
\paragraph{Effectiveness of DOS design}
To validate our design, we experiment with two degenerated versions of our framework: (1) only the canonical space, and (2) only the posed space without deformation field. 
The results, showed in \cref{fig:ablation_2}, clearly demonstrate that neither of these degenerated designs can match the performance achieved by our DOS design.

\vspace{-0.5em}
\section{Conclusions}
\vspace{-0.5em}
In this paper, we have introduced DreamAvatar, an effective framework for text-and-shape guided 3D human avatar generation. In DreamAvatar, we propose to leverage the parametric SMPL model to provide shape prior, guiding the generation with a rough shape and pose. We also propose a dual-observation-space design, facilitating the generation of more complete textures and geometry faithful to the target pose. 
Additionally, we propose to jointly optimize the loss computed from the full body and from the zoomed-in 3D head, effectively alleviating the multi-face ``Janus'' problem and improving facial details in the generated avatars.
Extensive experiments show that our method has achieved state-of-the-art 3D human avatar generation.

\vspace{-1.5em}
\paragraph{Limitations}
Despite establishing a new state-of-the-art, DreamAvatar encounters limitations: \textbf{(1)} Animation was not considered in our current implementation of DreamAvatar; \textbf{(2)} The model inherits biases from the pretrained diffusion model due to the text-image data distribution, such that performance on more frequently appeared subjects in the pretraining data may be better than the others.  

\vspace{-1.5em}
\paragraph{Societal impact}
Advancements in 3D avatar generation can streamline metaverse development. However, dangers exist regarding the nefarious use of this technology to generate plausible renderings of individuals. We encourage that research and usage to be conducted in an open and transparent manner.

{
    \small
    \bibliographystyle{ieeenat_fullname}
    \bibliography{egbib}
}


\end{document}